\pdfoutput=1
\documentclass[11pt]{article}

\usepackage[]{acl}

\usepackage{times}
\usepackage{latexsym}

\usepackage{acronym}
\usepackage{adjustbox}
\usepackage{bm}
\usepackage{booktabs}
\usepackage{numprint}
\usepackage{xspace}
\usepackage{multirow}

\acrodef{CIS}{conversational information seeking}
\acrodef{IR}{information retrieval}
\acrodef{NLP}{natural language processing}

\acrodef{LLM}{large language model}
\acrodefplural{LLM}{large language models}

\newcommand{\ourMethod}[0]{\textit{DAUS}\xspace}
\newcommand{\internalData}[0]{\textit{AutomotiveData}\xspace}

\usepackage[T1]{fontenc}
\usepackage[utf8]{inputenc}

\usepackage{microtype}

\title{Reliable LLM-based User Simulator for Task-Oriented Dialogue Systems}%: Evaluation and Error Detection}

\author{
\bf Ivan Sekuli\'c,
Silvia Terragni,
Victor Guimarães, 
Nghia Khau, \\
\bf Bruna Guedes,
Modestas Filipavicius,
André Ferreira Manso,
Roland Mathis \\
Telepathy Labs GmbH \\
Zürich, Switzerland \\
\texttt{firstname.lastname@telepathy.ai}
}

\begin{document}
\maketitle
\begin{abstract}
In the realm of dialogue systems, user simulation techniques have emerged as a game-changer, redefining the evaluation and enhancement of task-oriented dialogue (TOD) systems. These methods are crucial for replicating real user interactions, enabling applications like synthetic data augmentation, error detection, and robust evaluation. However, existing approaches often rely on rigid rule-based methods or on annotated data.

This paper introduces \ourMethod, a Domain-Aware User Simulator. Leveraging large language models, we fine-tune \ourMethod on real examples of task-oriented dialogues. Results on two relevant benchmarks showcase significant improvements in terms of user goal fulfillment. Notably, we have observed that fine-tuning enhances the simulator's coherence with user goals, effectively mitigating hallucinations -- a major source of inconsistencies in simulator responses.
\end{abstract}

\section{Introduction}
The field of dialogue systems has seen a notable surge in the utilization of user simulation approaches, primarily for the evaluation and enhancement of conversational search systems~\cite{owoicho2023exploiting} and task-oriented dialogue (TOD) systems~\cite{terragni2023context}. User simulation plays a pivotal role in replicating the nuanced interactions of real users with these systems, enabling a wide range of applications such as synthetic data augmentation, error detection, and evaluation~\cite{wan-etal-2022-unified,sekulic2022evaluating,Li2022DialogSimulation, balog2023user, ji2022achieving}.
%User simulator's task is to interact with the system in a way that resembles real users~\cite{balog2023user}, thus enabling synthetic data augmentation~\cite{wan-etal-2022-unified}, error detection \todo{[cite]}, and evaluation~\cite{sekulic2022evaluating} \tocheck{Should we also mention training loops? In our case policies, maybe more general decision making or training the system, or is there a reason we don't mention this?}.

%The simulator engages in multi-turn interactions with these systems, ideally until the given conversation goal is fulfilled.

\begin{figure*}
    \centering
    \includegraphics[width=0.73\linewidth]{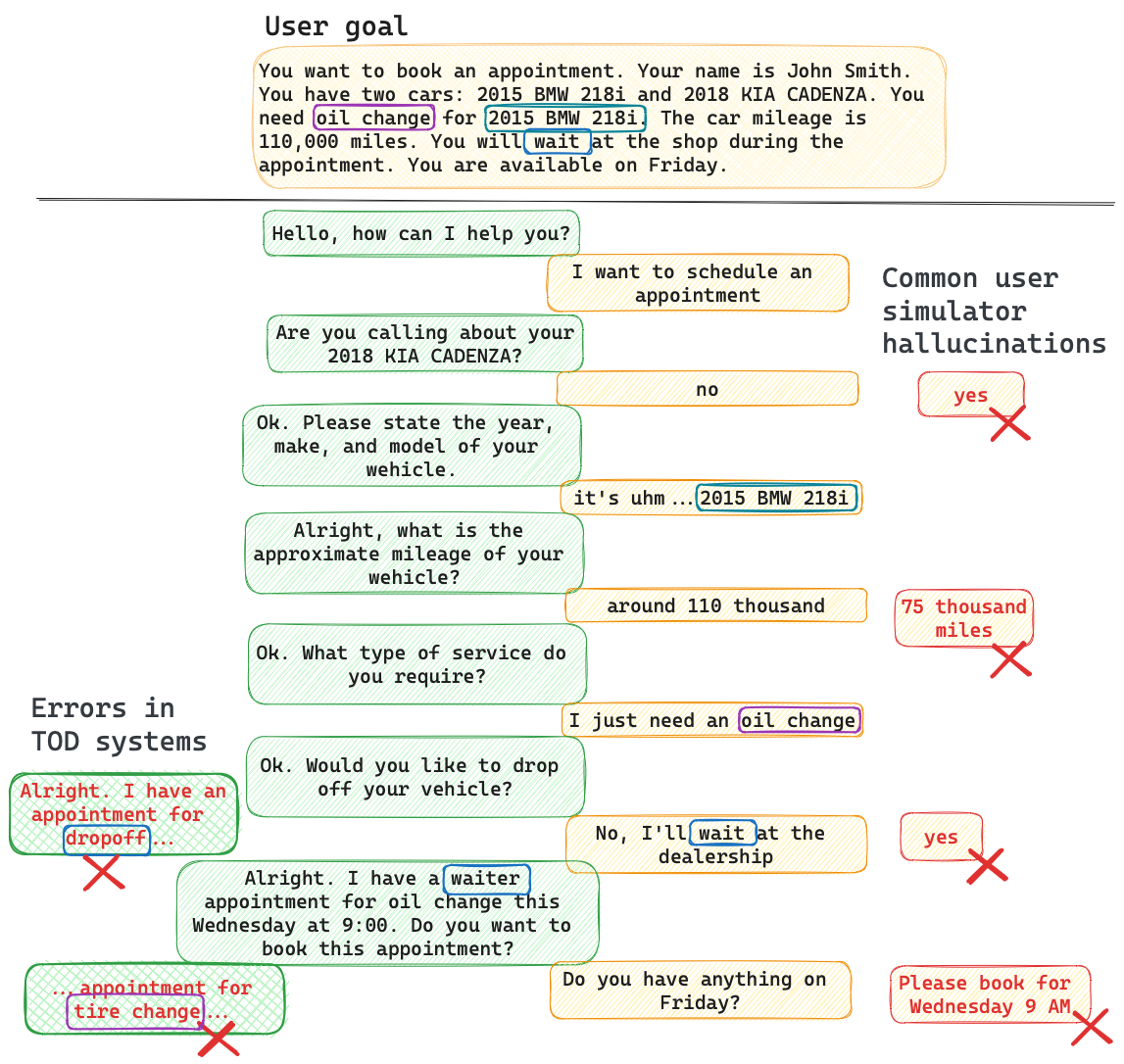}
    \caption{Example conversation between user simulator and TOD system. We aim to minimize common simulator's hallucinations (right) and thus ease the detection of TOD system failures (left).}
    \label{fig:intro_fig}
\end{figure*}

The significance of user simulation in the development and evaluation of dialogue systems is undeniable. However, the prevailing methodologies often rely on rudimentary rule- and template-based approaches, which can limit their adaptability and effectiveness~\cite{schatzmann2007agenda,Schatzmann2009agenda}. Furthermore, certain user simulation methods require a substantial amount of annotated data~\cite{Lin2021DomainindependentUS,Lin2022GenTUSSU,lin2023emous}, or a deep understanding of the internal workings of the dialogue system they interact with~\cite{schatzmann2007agenda, li2016user}.

The rise of generative capabilities of \acp{LLM} enabled user simulators to generate contextually appropriate responses in natural language, without the need for predefined rules~\cite{terragni2023context,davidson2023incontextUS}. 
This shift offers distinct advantages over traditional approaches:
% \tocheck{see slack channel discussion} : 
i) no human effort is needed to construct the rules; ii) it introduces lexical diversity into utterance generation to assess the robustness of downstream natural language understanding and enables testing of system's robustness to different dialogue paths.
However, \acp{LLM} are susceptible to hallucinations~\cite{hallucinations,terragni2023context}, resulting in inconsistency across dialogue turns or the generation of irrelevant information to the user's goal.
%\todoi{Mention limitations -- non-consistent across turns or only on utterance-level, faithfulness.}

In this paper, we introduce \ourMethod, a generative user simulator for TOD systems.
As depicted in Figure~\ref{fig:intro_fig}, once initialized with the user goal description,  \ourMethod engages with the system across multiple turns, providing information to fulfill the user's objectives.
Our aim is to minimize the commonly observed user simulator hallucinations and incorrect responses (right-hand side of Figure~\ref{fig:intro_fig}), with an ultimate objective of enabling detection of common errors in TOD systems (left-hand side of Figure~\ref{fig:intro_fig}).
Our approach is straightforward yet effective: we build upon the foundation of LLM-based simulators~\cite{terragni2023context, owoicho2023exploiting} and extend such approach by fine-tuning the LLM on in-domain dialogues, annotated with their user goals.
%We experiment with two data sources: internal user-system dialogues and the well-established MultiWoZ 2.1~\cite{eric2019multiwoz}.
%Our experiments demonstrate that this approach leads to more faithful dialogues, as the simulator is familiar with domain-specific terminology and strives to maintain the dialogue until all subtasks of the user's goal are accomplished. 
Notably, \ourMethod does not require insights into the inner-workings of the TOD system, its policy, nor system-specific functionalities, as it interacts with the TOD system strictly through natural language.

% We evaluate our approach by assessing whether all subtasks from the given goal were fulfilled and whether entities (e.g., user's personal information) are faithful, i.e., all of the generated utterances through the dialogue were grounded in the given user goal.
% We enable \ourMethod's interaction with two TOD systems: 1) internal TOD system for automotive domain; 2) ConvLab2 TOD system based on MultiWoZ data.
% Results show that \ourMethod significantly outperforms previous approaches based on goal fulfillment metrics, as well as faithfulness across relevant entities from the user goal.
% Specifically, we observe gains...

We summarize our contributions and findings as follows:
\begin{itemize}

    \item \textbf{Domain-Specific Adaptation}: \ourMethod fine-tunes a pre-trained LLM on domain-specific conversational data, enhancing the simulator's ability to maintain coherent and contextually relevant dialogues in a specific domain. %This adaptation enhances the simulator's ability to maintain coherent and contextually relevant dialogues in a specific domain, as it becomes familiar with domain-specific terminology. This adaptation contributes to more faithful and effective dialogues.

    \item \textbf{Reducing Simulator Hallucinations}: \ourMethod mitigates hallucinations originated from in-context learning approaches, which caused inconsistencies and irrelevant information in simulator responses. By fine-tuning on domain-specific data, our approach ensures more coherent and contextually relevant simulated dialogues.

    \item \textbf{Balancing Lexical Diversity in User Simulation}: \ourMethod employs LLMs for user simulation, offering a degree of lexical diversity in generated utterances. While not matching the diversity of in-context learning (partly due to hallucinations), it still provides language variety. %This level of diversity effectively evaluates downstream language understanding in dialogue systems and tests a wider range of TOD system capabilities.

    % \item We propose \ourMethod, capable of reliable and faithful multi-turn interaction with TOD systems. To this end, we show that fine-tuning on relevant data is beneficial.
    % \item We perform extensive qualitative analysis, focused on understanding the prevailing limitations of LLM-based user simulation.
    %\item We demonstrate a case-study for detecting errors in TOD systems, showcasing the potential of \ourMethod.
\end{itemize}

% \maybe{In intro:}
% \begin{itemize}
%     \item motivate user simulation
%     \item point out limitations: rule- and template-based, needs insight into inner-workings of the system
%     \item generative LLMs to the rescue -- we can generate utterances
%     \item some approaches do that on utterance level
%     \item recently, Silvia proposed to do it on conversation level --> given one goal, interact and evaluate in the end
%     \item problems have occurred, cite Silvia's
%     \item thus, we propose a new method, fine-tuned on in-domain data that outperforms recent state-of-the-art
% \end{itemize}

\section{Related Work}
%Our work is related to Task-Oriented Dialogue (TOD) systems and user simulation. In this section, we review relevant works on the topics.

\subsection{Task-Oriented Dialogue Systems}

The field of TOD systems, dedicated to interacting with users to accomplish specific tasks, has recently witnessed notable advancements~\cite{zhang2020recent}. Given the achievements of LLMs in various natural language processing tasks, there have been efforts to apply them to TOD systems~\cite{rColin2020,lOuyang2022}. A prominent application involves leveraging LLMs to extract users intents and entities, enhancing the Natural Language Understanding or Dialog State Tracking components~\cite{zhao2022description,gupta-etal-2022-show,madotto2021few,Madotto2020LanguageMA}.

%On one hand, \cite{zhao2022description} applies a sequence-to-sequence model to extract users' intent and entities, given a natural language description of the domain and the user's utterances.
%On the other hand, \cite{gupta-etal-2022-show} proposes to give labeled examples to the LLM, instead of a natural language description of the domain. The authors show that one-shot example with good examples is beneficial for TOD systems.

Furthermore, \citet{hudevcek2023llms} suggest that LLMs have the potential to be used off-the-shelf in TOD systems, even without fine-tuning for the specific TOD task, but their performance still lags behind supervised approaches. In response, an alternative approach underscores the benefits of fine-tuning specifically for TOD systems~\cite{bang2023task,hosseini2020simple,gupta-etal-2022-instructdial}. This line of research reveals that fine-tune LLMs can play a crucial role in enhancing the capabilities of TOD systems.
%\cite{hosseini2020simple} shows that fine-tuned GPT-2 model improves the performance of TOD systems.
%In addition, \cite{bang2023task} shows that fine-tuning different adapters for NLU, DST, NLG modules is beneficial to TOD systems.

\subsection{User Simulation}
The state of the art in user simulation for TOD systems has evolved significantly in the recent years. Initially, \citet{Eckert1997bigramUS} proposed the Bigram model, which estimates a user action conditioned on the system actions. Although efficient, this model does not account for the user goal coherence. Rule-based methods like Agenda-based~\cite{schatzmann2007agenda, Schatzmann2009agenda, Keizer2010agenda} addresses the coherence issue but relies on the manual definition of rules. 

Data-driven approaches, leveraging deep learning models~\cite{Gur2018rnnUS, Asri2016seq2seqUS,Lin2021DomainindependentUS,Lin2022GenTUSSU,lin2023emous}, overcome the rule-based constraints but require significant computational resources and annotated data. These methods mandate dialog annotation for user goal fulfillment at each turn. In-context learning approaches~\cite{terragni2023context} have recently gained traction, designing prompts using snippets of example dialogues, the user's goal (expressed in natural language as in~\citet{terragni2023context}, or structured format as in~\citet{davidson2023incontextUS}), and the dialog history. While these approaches demand fewer resources than fine-tuning methods and eschew manual annotation, they underscore limitations of LLMs, including hallucinations, repetitions, and incomplete user goal fulfillment.

%DIALOGIC: Controllable Dialogue Simulation with In-Context Learning \cite{li-etal-2022-controllable}.

%CPed: "Agenda-based simulation is an example that uses a stack of actions to simulate the user’s goals (Schatzmann et al., 2007; Schatzmann and Young, 2009; Keizer et al., 2010). These models can generate realistic user behavior, but they require manual rule definition and are not transferable to new domains."
\section{Generative User Simulator}
\label{sec:methodology}
In this section, we define the task of generative user simulation for TOD systems.
Moreover, we describe our approach, based on fine-tuned \acp{LLM}. 

\subsection{Background}
\label{sec:methodology:task}
When interacting with a TOD system, users aim to fulfill their goal, e.g., book a flight, or cancel their reservation in a restaurant.
Therefore, a user simulator ($U$), designed to imitate a real user, interacts with the TOD system ($S$) with a given user goal $\mathcal{G}$.
Formally, interactions are a sequence of utterances, where the system's utterances $s$ and the user's utterances $u$ take turns, forming a dialogue history $\mathcal{H} = [s_1, u_1, \dots, s_{t}, u_{t},  \dots, s_N, u_N]$, with $s_t$ and $u_t$ corresponding to system's and user's utterance at turn $t$, respectively, and $N$ being the total number of exchanged utterances.

We define the user goal $\mathcal{G}$ as all the information the user requires to achieve their aim. An example of user goal is the following: \textit{You want to try an Indian restaurant. The restaurant must be cheap and in the center. Book a table for 2 people at 8PM}.
At the end of the dialogue, we expect the user simulator to have fulfilled $\mathcal{G}$. 
%This includes the user's subtask (e.g., to book an appointment), personal information about the user (e.g., name, phone number), and user's availability (e.g., available on Mondays). 
While the goal $\mathcal{G}$ can be represented either in structural format~\cite{davidson2023incontextUS} or in natural language~\cite{terragni2023context}, in this work we focus on $\mathcal{G}$ represented in natural language. $\mathcal{G}$ is usually defined by a domain expert or randomly sampled.

%The interaction is done in a conversational manner, by each $U$ and $S$ responding to each other in natural language, turn by turn.
% At the end of the dialogue, the fulfillment of $\mathcal{G}$ is assessed depending on the type of TOD system we are interacting with, in relation to $\mathcal{H}$.
% In general, we care about two main streams of evaluation approaches: 1) whether the given subtask in $\mathcal{G}$ was successfully fulfilled (e.g., was the correct flight booked); and 2) whether all of the utterances $u_t$ have been faithful to $\mathcal{G}$ (e.g., personal information provided to $S$ is correct).
% We detail our evaluation procedure in Section~\ref{sec:evaluation}.

\subsection{Our Approach}
\label{sec:methodology:ours}

We propose Domain-Aware User Simulator (DAUS), a model that relies on learning the specifics of interactions with a TOD system from conversational data. 
The data needs to contain the goal $\mathcal{G}$ and the dialogue history $\mathcal{H}$.
Typically, such datasets are derived from user conversations with production TOD systems, or created and curated through crowd-sourcing or user studies.
%In this paper, as detailed in Section~\ref{sec:setting:data}, we experiment with two data sources, one internal and one public.

We cast the above-described problem of simulator's goal fulfillment to an utterance-level generation task.
Specifically, the main task of $U$ is to generate the next utterance $u_t$ by modeling:
\begin{equation}
\label{eq:task}
    u_t = \phi(\mathcal{G}, \mathcal{H})
\end{equation}
where $\phi$ is the function to generate a user utterance.
The $u_t$ needs be aligned with $\mathcal{G}$ and $\mathcal{H}$, i.e., it needs to be faithful towards the given goal, as well as coherent with the dialogue so far.

Given that both $\mathcal{G}$ and $\mathcal{H}$ are in natural language, we model $\phi$ from Eq.~\ref{eq:task} with a language modeling-based approach.
Specifically, we first construct a prompt to feed an LLM, by combining $\mathcal{G}$ and $\mathcal{H}$.
We further employ the LLM to generate the $u_t$ in auto-regressive fashion:
% \begin{equation}
%     p_{LLM}(x) = LLM(Prompt(\mathcal{G}, \mathcal{H}, s_t)
% \end{equation}
\begin{equation}
\label{eq:generation}
p_{LLM}(\boldsymbol{u_t}|\mathcal{G},\mathcal{H}) = \prod_{i=1}^{n}p_{LLM}(x_t^i|x_t^{<i},\mathcal{G},\mathcal{H})
\end{equation}
where $x_t^i$ is the $i$-th token of the utterance at turn $t$.
% Our approach consists of fine-tuning the LLM on in-domain data, by setting the $u_t$ as a language modeling target.
% Thus, the model learns to generate user utterances aligned with the given goal $\mathcal{G}$.
% The in-domain data contains dialogues between real users, with real goals $G$, and TOD system(s).
We break down the dialogue from the data by turn, yielding $N$ data points for each conversation.
%The details on fine-tuning specifics of the LLMs and the data used are in Section~\ref{sec:setting:finetuning} and Section~\ref{sec:setting:data}, respectively.

Regarding the interaction between our fine-tuned LLM and a TOD system, we follow the same paradigm from \citet{terragni2023context}. \ourMethod receives a fresh prompt, which comprises the user's goal for the ongoing dialogue and the cumulative dialogue history. Unlike \citet{terragni2023context}, we do not provide any example dialogues to serve as shots. We additionally post-process the generated utterance to ensure that a clean message is passed to TOD systems (i.e., removal of special characters and trailing tokens). 

%These dialogues should follow the same structure of the data on which the model has been fine-tuned. For more details on the architecture, 

\section{Experimental Setting}
\label{sec:setting}
% experimental setting
In this section, we describe datasets, implementation details, and experimental setting for simulator-system interactions.

\subsection{Data sources}
\label{sec:setting:data}

\begin{table}[ht]
    \centering
     \caption{Dataset Statistics (after pre-processing).}
    \label{tab:dataset_stats}
    \resizebox{\linewidth}{!}{  
\begin{tabular}{@{}crrr@{}}
\toprule
Dataset & 
  \begin{tabular}[c]{@{}c@{}}Avg \\\# Turns\end{tabular} &
  \begin{tabular}[c]{@{}c@{}}Avg \# Words \\per User\\Utterance\end{tabular} &
  \begin{tabular}[c]{@{}c@{}}Avg \# Words \\ per TOD\\Utterance\end{tabular} \\ \midrule
MultiWOZ &
  5.86 &
  13.13 &
  14.86 \\
\internalData &
  11.20 &
  3.44 &
  12.06 \\ \bottomrule
\end{tabular}
    }
   
\end{table}

We consider two data sources to evaluate our approach.
First, we experiment on internal dialogue data of user-TOD system phone call interactions within the automotive industry, dubbed \internalData.\footnote{In order to protect our users' privacy, we do not release any user data nor models fine-tuned on user's data. Examples presented throughout the paper are synthetically constructed, whilst preserving realistic user goals. Users have been informed about and have consented to data collection.}
Second, we use the well-established dataset of multi-domain TOD systems -- MultiWOZ~2.1~\cite{eric2019multiwoz}.
Both data sources contain user goal $\mathcal{G}$ in natural language and multi-turn dialogues (compliant with Section~\ref{sec:methodology:task}). 
%We break down the dialogues into utterance-level data points, thus constructing the dialog history $\mathcal{H}$, the most recent system utterance $s_t$, and the target utterance $u_t$. 
For each dataset, we randomly sample 2,500 dialogues for training, 300 for testing and 300 for validation. The statistics of the resulting datasets are reported in Table~\ref{tab:dataset_stats}.

\subsection{TOD Systems}
\label{sec:setting:TOD}
\ourMethod communicates with TOD systems through natural language, making it system-agnostic.  For our user simulator fine-tuned on \internalData, we employ an internal TOD system. To evaluate \ourMethod fine-tuned on MultiWOZ, we use the ConvLab2 framework~\cite{zhu2020convlab2}, extended by \citet{terragni2023context}, which integrates LLM-based few-shot user simulators.\footnote{ \url{https://github.com/telepathylabsai/prompt-based-user-simulator}.}
We use the same TOD the authors used in their original work. We identify a challenge with the default stopping criteria that prematurely end dialogues when users express gratitude. This does not always signify the end of the interaction as users may continue with their goals (e.g., \emph{``Thanks for booking my flight. I also need a hotel''} would terminate the conversation).
Therefore, we modify the criteria to exclude termination on \emph{``thanks''} intent. 
We consequently re-run the experiments presented in \citet{terragni2023context}.
Moreover, we publicly release the updated framework and the user simulator fine-tuned on MultiWOZ~2.1 at \url{https://github.com/telepathylabsai/finetuned-user-simulator}.

%the user simulator has to indicate the termination of a conversation. 
%In \citet{terragni2023context}, they terminate dialogues if the user's goal is complete (not necessarily fulfilled) or if the user expressed gratitude ("thanks" intent) or bid farewell ("bye" intent). 
%However, we observed that using "thanks" as a termination signal resulted in numerous unfinished conversations, prompting us to remove this criterion.

\subsection{User Goal Settings}
\label{sec:setting:user_goals}
For the MultiWOZ data within ConvLab2 framework, we follow the previous work for construction of the user goals~\cite{zhu2020convlab2, terragni2023context}.
Specifically, the user goals are randomly sampled, conditioned on the domains and entities frequency in the training data. We generate $100$ dialogues per user goal.

For evaluation on our internal TOD system, a domain expert manually defined user goals for 8 test cases, detailed in Appendix~\ref{app:task}.
The test cases vary depending on the complexity and the main task that the simulator has to fulfill. 
As such, we label the test cases accordingly: \emph{B} for \emph{book} appointment task, \emph{C} for \emph{cancel} appointment task, \emph{R} for \emph{reschedule} appointment task.
Moreover, each label is associated with a graded difficulty indicator, i.e., \emph{easy} or \emph{hard}. We generate $100$ dialogues per test case (i.e., per user goal).

\subsection{Fine-tuning Details}
\label{sec:setting:finetuning}
We conduct our experiments with the recently released open-source LLM --- \texttt{Llama-2}~\cite{touvron2023llama}.
The prompt, mentioned in Section~\ref{sec:methodology:ours}, is constructed by concatenating the task description, user goal $\mathcal{G}$, and the dialog history $\mathcal{H}$.
Moreover, we separate every utterance with a special ``\textit{<endturn>}'' token.

We utilize LoRA~\cite{hu2021lora} -- a parameter-efficient fine-tuning technique, capable of reaching performances comparable to fully fine-tuned models, whilst requiring only a fraction of the computational resources. 
% explain more?
We adhere to the hyperparameter recommendations and instructions of the recent work on the topic~\cite{hu2021lora, he2021towards} and use the following LoRA hyperparameters throughout the experiments: rank $r$ of $64$, $\alpha$ of $32$, and dropout of $0.05$.
Moreover, we optimize attention layers (query and key matrices) of the \texttt{Llama-2} model.
We use the 13B \texttt{Llama-2} version for the main experiments, and the 7B version for comparison and the generalization study.
We perform hyperparameter grid search for learning rate on the dev sets of our datasets. We settle for $lr=3e^{-5}$ and the batch size of 12 and 32 for the 13B and 7B versions, respectively.

% and do not perform extensive hyperparameter search, as it falls out of scope of this paper.

\subsection{Baselines}
\label{sec:setting:baselines}
We compare our Llama-2 fine-tuned model with several pre-trained models in zero-shot or few-shot fashion, following \cite{terragni2023context,davidson2023incontextUS}. In particular, we consider the following pre-trained models:
\begin{itemize}\itemsep0.2em 
    \vspace{-2mm}\item Llama 2 with 13B parameters.
    \item GPT-3.5 Turbo4 (Chat-GPT), version 0613~\cite{brown2020language}. For data privacy reasons, we employ this model only for the MultiWOZ experiments.
    \item Flan-T5~\cite{chung2022flant5} with 3B parameters (XL), to reproduce  results of \citet{terragni2023context}.
\end{itemize}
In addition to the LLM-based models, we consider an agenda-based simulator (ABUS)~\cite{wen2015nlg}, designed specifically for MultiWOZ within ConvLab2 framework, thus requiring the knowledge of TOD system's policy. We include two variants of \texttt{ABUS}: the first with template-based NLG and the second with data-driven NLG, dubbed \texttt{ABUS-T} and \texttt{ABUS-D}, respectively. Let us notice that ABUS is a strong baseline, as it is tailored for communicating with the MultiWOZ-based TOD from ConvLab2, therefore it is included as a reference of the potential upper-bound for user goal fulfillment performance.
We follow \citet{terragni2023context} and set the temperature for inference to $0.9$ for all MultiWOZ experiments, and $0.7$ for internal experiments (value chosen through grid search).

% We evaluate \ourMethod against recent state-of-the-art approaches in the area of user simulation. We also adapt those methods to the most recent \acp{LLM}.
% \noindent \textbf{Zero-shot Llama-2} relies on the state-of-the-art open-source \acp{LLM} for generating the $u_t$.

% \noindent \textbf{Few-shot FlanT5}~\cite{terragni2023context} uses in-context learning with instruction-tuned T5, Flan-T5~\cite{chung2022flant5}, in a few-shot scenario for goal fulfillment task.

% \noindent \textbf{Few-shot Llama-2} extends \citet{terragni2023context} with the most recent open-source \acp{LLM}.

% \noindent \textbf{GPT-3.5}~\cite{brown2020language} in a 2-shot setting. We use OpenAI GPT-3.5 Turbo4 (Chat-GPT), version 0613.\footnote{We employ this model only for the MultiWOZ experiments.}

% \noindent \textbf{ABUS}~\cite{wen2015nlg} is an agenda-based user simulator model designed specifically for MultiWOZ within ConvLab2 framework, thus requiring the knowledge of TOD system's policy. We include two variants of \texttt{ABUS}: the first with template-based NLG and the second with data-driven NLG, dubbed \texttt{ABUS-T} and \texttt{ABUS-D}, respectively. Agenda-based user simulator is included as a reference of the potential upper-bound for user goal fulfillment performance.

\section{Evaluation}
\label{sec:evaluation}
We comprehensively evaluate our method, aiming to assess its ability to achieve designated user goals in dialogues and its impact on lexical diversity when aligning with real user language patterns. 
Moreover, we perform qualitative analysis of simulated dialogues via human evaluation.
In this section, we detail these evaluation procedures.

Additionally, we examine utterance-level metrics, comparing generated utterances with those in the target dataset, using both general natural language generation and domain-specific entity-based metrics. However, we found that these metrics poorly correlate with the simulator's task completion. Detailed information about these metrics and their results can be found in Appendix~\ref{app:utterance_level_metrics}.

%We show the goal fulfillment and diversity results in Section~\ref{sec:results:goal}.

%Additionally, we carry out a qualitative analysis of the generated dialogues and identify prevalent errors in the LLM-based simulators, which we elaborate in Section~\cite{sec:analysis:qualitative}.
\subsection{Goal Fulfillment Evaluation Metrics}
\label{sec:eval:goal_fulfillment_metrics}

Our objective is to evaluate the goal fulfillment at the end of the dialogue.
For MultiWOZ experiments, we consider well-known metrics such as Success, Completion and Book rate.
These metrics aim to capture how successful was the dialogue in terms of fulfilling specific subtasks from the user goal (e.g., whether the restaurant is booked).
We also compute the average precision ($P$), recall ($R$) and $F_1$ scores by matching the entities expressed through the simulated dialogue to the ones in the initial user goal. 
These metrics aim to assess the simulator's faithfulness and consistency of entities with the user goal (e.g., whether the correct restaurant \emph{type} was booked).
%Additionally, we show the average number of dialog turns and successful dialog turns. 
For a comprehensive understanding of the metric definitions, please refer to \citet{zhu2020convlab2} and \citet{terragni2023context}. 

Regarding our in-house TOD, it is worth noting that we do not differentiate between \textit{book}, \textit{inform} and \textit{request} entities. Therefore, we adapt the mentioned metrics, except for the Book Rate, while considering all entities as inform entities.
Moreover, we compute several metrics specific to automotive domain: \emph{user subtask} indicating whether the subtask (\emph{book}, \emph{cancel}, or \emph{reschedule} the appointment) matches the one given in the user goal; \emph{caller info} and \emph{car info} indicating whether user information (name, phone number) and vehicle information (car year, make, and model) match the ones in the goal, respectively; \emph{transport type} assessing the chosen transport type (e.g., dropping of the vehicle, waiting for the service in the dealership).

% we consider the following goal fulfillment metrics:   
% \begin{itemize}
%     \item \textbf{Succ Rate} evaluates matches for booked and informed slots. It is 1 if all the values match, 0 otherwise.
%     \item \textbf{Compl Rate} is 1 if \textbf{all} the booked slots in the user goal have been filled (disregarding if the value is correct), 0 otherwise.
%       \item \textbf{Book Rate} is the ratio of booked slots having a value matching the one in the user goal. We report this metric for MultiWOZ only, because it makes a distinction between \textbf{book} and \textit{inform} entities. 
%     \item \textbf{Inform Prec/Rec/F1} assess if the system provides the requested information accurately. True Positives (TP) are slots correctly mentioned, False Positives (FP) are slots with invalid values or only user-mentioned, and False Negatives (FN) are slots mentioned by the system with valid values but not in the goal.
%     \item \textbf{DT} and \textbf{Successful DT} count the number of turns per dialog and per successful dialog respectively. 
% \end{itemize}

\subsection{Lexical Diversity of Generated Utterances}
Lexical diversity (LD) is a measure of word variability and vocabulary size of a given text corpus, in our case, the set of generated user utterances from 100 conversations. We report MTLD scores \cite{mccarthy2005assessment}, and a number of unigram words (Unig) and average user utterance length (UttLen). 
LD results are reported in Section~\ref{sec:results:diversity}.
%LD baseline for both datasets are produced by sampling 100 conversations from each training dataset and calculating LD metrics. This process is repeated 1,000 times.

\begin{table*}[ht]
\centering
\caption{Results of goal fulfillment task in simulator interaction with the internal TOD system. The results are averaged across the eight user goals. %The $\dagger$ and $\ddagger$ signs indicate statistically significant difference compared to \texttt{Llama-2 2-shot} and \texttt{FlanT5-XL} baselines, respectively. The significance is reported under two-sided t-test with $p < 0.01$ with Bonferroni correction for multiple comparisons.
}% Lexical diversity metrics calculated from US utterances from 100 conversations for each of 8 tasks, and averaged.}% \todoi{TODO somehow add a new row with lex diversity metrics for actual user utterances from the training dataset: avgL 2.0, unigrams 256, mtld 20.5.}}
\label{tbl:internal_results}
\adjustbox{max width=\textwidth}{%
\begin{tabular}{lcrrrrrrrrrrrrr}
\toprule
\multicolumn{2}{l}{Model} & \begin{tabular}[c]{@{}l@{}}Num\\ Shots\end{tabular}& \begin{tabular}[c]{@{}l@{}}Compl\\ Rate\end{tabular} & \begin{tabular}[c]{@{}l@{}}Succ\\ Rate\end{tabular} & $P$ & $R$ & $F_1$ & \begin{tabular}[c]{@{}l@{}}User\\ Subtask\end{tabular} & \begin{tabular}[c]{@{}l@{}}Caller\\ Info\end{tabular} & \begin{tabular}[c]{@{}l@{}}Car\\ Info\end{tabular} &  \begin{tabular}[c]{@{}l@{}}Transport\\ Type\end{tabular} & UttLen & Unig & MTLD\\
\midrule
\multicolumn{2}{l}{\begin{tabular}[c]{@{}l@{}}\texttt{FlanT5-XL} \\ \cite{terragni2023context}\end{tabular}} & 2 & 0.46 & 0.27 & 0.72 & 0.86 & 0.76 & 70.9 & 85.5 & 65.6 & 39.2 & \textbf{2.8} & \textbf{209} & \textbf{23.4} \\
\midrule
% 0-shot & 50.4 & 34.5 & 13.0 & 0.62 & 0.89 & 0.69 & 88.8 & 72.2 & 26.3 & 12.8 \\
% 1-shot & 65.6 & 37.4 & 12.0 & 0.67 & 0.89 & 0.74 & 89.1 & 81.6 & 33.0 & 8.0 \\
% 2-shot & 68.9 & 35.8 & 14.8 & 0.66 & 0.91 & 0.74 & 90.3 & 80.2 & 28.8 & 8.0 \\

\multirow{3}{*}{\texttt{Llama-2}}& & 0 & 0.35 & 0.13 & 0.62 & 0.87 & 0.69 & 50.4 & 88.8 & 72.2 & 12.8 & 2.4 & 161 & 15.5 \\
 && 1 & 0.37 & 0.12 & 0.67 & 0.89 & 0.74 & 65.6 & 89.1 & 81.6 & 8.0 & 2.0 & 149 & 14.5 \\
 & &2 & 0.36 & 0.15 & 0.66 & 0.91 & 0.74 & 68.9 & 90.3 & 80.2 & 8.0 & 2.0 & 129 & 13.7 \\
 \midrule
\multicolumn{2}{l}{\ourMethod} & 0 & \textbf{0.51} & \textbf{0.40} & \textbf{0.91} & \textbf{0.92} & \textbf{0.91} & \textbf{99.5} & \textbf{98.5} & \textbf{99.0}& \textbf{80.7}& 1.7 & 112 & 16.5 
\\ \bottomrule
\end{tabular}
}
\end{table*}

\subsection{Qualitative Analysis}
\label{sec:evaluation:qualitative}
During the analysis of the generated simulated dialogues, we observed several re-occurring issues.
We categorize them as the simulator's failure (\emph{hallucination}, \emph{incomplete user goal} fulfillment, or \emph{looping/repeating} utterances across turns) or TOD system's failure (\emph{NLU misclassification} due to missing user's intent or entities, \emph{forcing end of dialogue}, or \emph{looping/repeating} utterances).
Our aim is to assess the prevalence of these patterns and identify potential limitations of LLM-based user simulators.
To this end, we employ three annotators to annotate $45$ dialogues generated with an LLM-based baseline and $45$ dialogues generated with \ourMethod within ConvLab2 framework.
The annotators are domain-experts and employees of the authors' institution.
We provide guidelines for each of the categories and go through an on-boarding process with the annotators.
The labels for each of the dialogues are determined by majority vote.
Annotators reach moderate to good agreement, as measured by Fleiss' $\kappa$, detailed in Appendix~\ref{app:kappa}.
%The findings of our qualitative analysis are discussed in Section~\ref{sec:analysis:qualitative}.

\section{Results}
\label{sec:results}

In this section, we examine our study's findings across three main threads. 
%\begin{itemize}
    First, we investigate the impact of fine-tuning LLMs with domain-specific data on goal fulfillment in dialog interactions (Section~\ref{sec:results:goal}).
    Next, we explore the link between fine-tuning and the lexical diversity of generated utterances (Section~\ref{sec:results:diversity}).
    Finally, we assess whether the adaptability of LLM-based user simulators to unseen user tasks is influenced by the diversity of subtask types in their training data (Section~\ref{sec:analysis:generalization}).
%\end{itemize}

%In this section, we report our results with aim to answer the following three research questions:

%\noindent\textbf{RQ1:} Does fine-tuning on domain data lead to more reliable LLM-based user simulators?% We address this question in Sections~\ref{sec:results:goal} and \ref{sec:results:multiwoz}.

%\noindent\textbf{RQ2:} Does fine-tuning have an impact on the lexical diversity of the generated utterances?% We discuss this question in Section~\ref{sec:results:diversity}.

%\noindent\textbf{RQ3:} Is our fine-tuned model able to generalize to unseen user goals? 
%we report the results and discuss the findings of LLM-based user simulators' interactions with the aforementioned TOD systems.

%\subsection{Goal Fulfillment}

\subsection{Goal Fulfillment} 

\label{sec:results:goal}

\paragraph{Internal TOD System.}

% wrong order of magintude
% \multirow{3}{*}{\texttt{Llama-2}} & 0-shot & 50.4 & 0.35 & 0.13 & 0.62 & 0.87 & 0.69 & 88.8 & 72.2 & 26.3 & 12.8 & 2.4 & 161 & 15.5 \\
%  & 1-shot & 65.6 & 0.37 & 0.12 & 0.67 & 0.89 & 0.74 & 89.1 & 81.6 & 33.0 & 8.0 & 2.0 & 149 & 14.5 \\
%  & 2-shot & 68.9 & 0.36 & 0.15 & 0.66 & 0.91 & 0.74 & 90.3 & 80.2 & 28.8 & 8.0 & 2.0 & 129 & 13.7 \\
%  \midrule
% \multicolumn{2}{l}{\ourMethod} & \textbf{99.5}^{\dagger\ddagger}$ & \textbf{0.51}$ & \textbf{0.40}$ & \textbf{0.91}^{\dagger\ddagger}$ & \textbf{0.92}$ & \textbf{0.91}^{\dagger\ddagger}$ & \textbf{98.5}$ & \textbf{99.0}^{\dagger\ddagger}$ & \textbf{42.8}$ & \textbf{80.7}^\dagger$ & 1.7 & 112 & 16.5 
% \\ \bottomrule

% \begin{tabular}{lllllll}
% \toprule
% Model & \begin{tabular}[c]{@{}l@{}}User\\ Task\end{tabular} & \begin{tabular}[c]{@{}l@{}}Compl\\ Rate\end{tabular} & \begin{tabular}[c]{@{}l@{}}Inform\\ $F_1$\end{tabular} & Car Info \\
% \midrule
% \begin{tabular}[c]{@{}l@{}}FLAN-T5\\ JS \cite{terragni2023context}\end{tabular} & \\
% \begin{tabular}[c]{@{}l@{}}Llama-2-13B\\ 0-shot\end{tabular} & 50.3 & 34.5 & 0.69 & 72.1 \\

% \begin{tabular}[c]{@{}l@{}}Llama-2-13B\\ 2-shot\end{tabular} & 68.8 & 35.8 & 0.74 & 80.0 \\
% \ourMethod & 99.5 & 51.3 & 0.91 & 0.99

\begin{table*}[]
\centering
\caption{Performance on MultiWOZ 2.1 within ConvLab2 framework.}
\label{tbl:multiwoz}
\small
%\adjustbox{max width=\columnwidth}{%
\begin{tabular}{lcrrrrrrrrrr}
\toprule
\multicolumn{2}{l}{Model} & \begin{tabular}[c]{@{}l@{}}Num\\Shots\end{tabular} &\begin{tabular}[c]{@{}l@{}}Compl\\ Rate\end{tabular} & \begin{tabular}[c]{@{}l@{}}Succ\\ Rate\end{tabular} & \begin{tabular}[c]{@{}l@{}}Book\\ Rate\end{tabular} & $P$ & $R$ & $F_1$  & UttLen & Unig & MTLD \\
\midrule
%\multicolumn{2}{l}{\texttt{Reference train}} & x & x & x & x & x & x & 12.6 & 683 & 61.0 \\
\multicolumn{2}{l}{\begin{tabular}[c]{@{}l@{}}\texttt{ABUS-T}\\ \cite{wen2015nlg}\end{tabular}} & - & 0.93 & 0.83 & 0.85 & 0.84 & 0.94 & 0.86 & \textbf{17.4} & 527 & 46.9 \\
\multicolumn{2}{l}{\begin{tabular}[c]{@{}l@{}}\texttt{ABUS-D}\\ \cite{wen2015nlg}\end{tabular}} & - & 0.86 & 0.60 & 0.75 & 0.87 & 0.90 & 0.87 & 9.8 & 327 & 28.0 \\
\midrule
\multicolumn{2}{l}{\begin{tabular}[c]{@{}l@{}}\texttt{FlanT5-XL}\\ \cite{terragni2023context}\end{tabular}}  & 2 & 0.19 & 0.13 & 0.46 & 0.45 & 0.39 & 0.39 & 13.7 & \textbf{888} & 41.2 \\
\multirow{2}{*}{\texttt{Llama-2}}& & 0 & 0.07 & 0.04 & 0.13 & 0.31 & 0.21 & 0.23 & 8.1 & 697 & 30.7 \\
& &2& 0.09 & 0.08 & 0.30 & 0.46 & 0.34 & 0.39 & 10.0 & 765 & 38.8 \\
\multicolumn{2}{l}{\texttt{GPT-3.5}} & 2 & 0.35 & 0.19 & 0.34 & 0.49 & 0.52 & 0.48 & 16.3 & 626 & 38.1 \\\midrule
\multicolumn{2}{l}{\ourMethod} & 0 &\textbf{0.41} & \textbf{0.29} & \textbf{0.66} & \textbf{0.69} & \textbf{0.69} & \textbf{0.67} & 10.6 & 789 & \textbf{54.9}
\\ \bottomrule
\end{tabular}
%}
\end{table*}
Table~\ref{tbl:internal_results} shows results on the goal fulfillment task of \ourMethod and the baselines detailed in Section~\ref{sec:setting:baselines}, averaged across different user goals. We present the results per each of the eight specific user goals, detailed in Section~\ref{sec:setting:user_goals}, in Appendix~\ref{app:results_per_task} for space-saving purposes.

As a first remark, \ourMethod outperforms all of the baselines across all the goal fulfillment metrics.  We observe the largest improvements for domain-specific metrics, e.g., precision and recall of relevant entities and accuracy of the transport type. 
This indicates that fine-tuning on in-domain data improves simulator's knowledge of the domain-specific terminology. We further expand on this observation in Section~\ref{sec:analysis:qualitative}.
%Statistically significant difference in performance is observed for both goal fulfillment metrics (i.e., user subtask identification), as well as entity-based metrics (e.g., entity $F_1$, caller info, transport type).

Regarding the baselines, \texttt{FlanT5}, employing 2 shots as examples, is the second best model. As observed in~\cite{terragni2023context} as well, this instruction fine-tuned model outperforms \texttt{Llama-2} with 2 shots in most of the cases. 

%\texttt{Llama-2} performs better in a few-shot setting, a finding similar to \citet{terragni2023context}.

%Overall, results strongly support our approach of fine-tuning LLMs on domain-specific data.
%\todoi{Can we comment on these results here or not? I.e., some tasks are harder some easier?}

\paragraph{MultiWOZ Data within ConvLab2.}
\label{sec:results:multiwoz}
We show the goal fulfillment performance of \ourMethod and the baselines in interaction with ConvLab2's TOD system on MultiWOZ~2.1 in Table~\ref{tbl:multiwoz}. As in Section~\ref{sec:results:goal}, we observe strong performance of \ourMethod.
Specifically, \ourMethod outperforms all of the in-context learning approaches in terms of goal fulfillment, including prior state-of-the-art~\cite{terragni2023context}. 
Moreover, our method outperforms few-shot \texttt{GPT-3.5}, a model significantly larger than ours (estimated 175 billion parameters vs 13 billion).
This further suggests the benefits of fine-tuning LLMs on domain-specific conversational data, as stronger performance can be achieved with significantly smaller LLMs, thus reducing the computational requirements of the simulator. %, as well as reducing the carbon footprint.

As a general remark, results on both benchmarks, i.e., the ConvLab2 and our internal one, show significant improvements across multiple goal fulfillment metrics. Thus, we conclude that \ourMethod indeed does lead to more consistent, reliable, and faithful LLM-based user simulators. We will discuss these results more in depth in our qualitative analysis in Section~\ref{sec:analysis:qualitative}.

% \subsection{Scaling of LLMs}
% \label{sec:results:scaling}
% Table~\ref{tbl:results_scaling} shows performance of different model sizes. 
% Make conclusions about potential overall improvements in the goal fulfilment task if we used larger models, based on untrained approaches.

% \begin{table}[]
% \caption{\todoi{Results of different sizes of LLMs and impact on goal fulfilment performance. Consider few-shot scenario.}}
% \label{tbl:results_scaling}
% \begin{tabular}{@{}lllll@{}}
% \toprule
%           & GF & Metric2 & Metric3 & Metric4 \\ \midrule
% llama 7B  &    &         &         &         \\
% llama 13B &    &         &         &         \\
% llama 70B &    &         &         &         \\ \bottomrule
% \end{tabular}
% \end{table}

\subsection{Lexical Diversity}
\label{sec:results:diversity}
Lexical diversity (LD) of generated user utterances from internal TOD system and MultiWOZ experiments is presented in the last 3 columns of Tables~\ref{tbl:internal_results} and \ref{tbl:multiwoz}.
We observe a drop in LD, as measured by the length of the generated utterances and the total number of unigrams, when \ourMethod is fine-tuned on \internalData. This suggests a limited vocabulary in the training data, which is expected due to the real users often responding with one or two words, especially in the \emph{cancel} task. \ourMethod had a relatively high MTLD score, because of the correctly generated caller, car and transport entities, which usually have unique values. However, a low unigram score is due to averaging metrics over 8 user tasks, where only 3 of them are the entity-rich \emph{book} task. Meanwhile, the higher LD of \texttt{FlanT5}-based method is due to its prevalent hallucinations, thus falsely inflating the LD scores by generating out-of-context content (see Section~\ref{sec:analysis:qualitative}).

In MultiWOZ-based experiments, results indicate higher LD than ABUS baselines, as measured by MTLD, while the generated utterances are slightly shorter compared to \texttt{FlanT5}.
As such, \ourMethod does not seem to lose LD during fine-tuning on MultiWOZ, while fine-tuning on \internalData seems to reduce it slightly. 
This can be explained by the fact that \internalData contains both specific vocabulary and utterances from real product users, which makes it hard for in-context learning approaches to imitate.
On the other hand, fine-tuning procedure enables the model to learn the particulars of such interactions.

\subsection{Generalization to Unseen User Tasks}
\label{sec:analysis:generalization}

Table~\ref{tbl:user_tasks} shows the percentage of successful subtask identifications for four variants of our model: \ourMethod fine-tuned on the full dataset described in Section~\ref{sec:setting:data}, and \ourMethod fine-tuned on modified datasets by removing certain subtasks (\emph{book (B)}, \emph{cancel (C)}, or \emph{reschedule (R)}) from the training sets.
With this experiment, we aim to assess the generalization abilities of our approach.

\begin{table}[h]
\caption{Percentage of dialogues with successfully identified subtask types across the test cases, with models fine-tuned on specific combinations of subtask types.}
\label{tbl:user_tasks}
\centering
%\adjustbox{width=0.66\columnwidth}{%
\small
\begin{tabular}{lrrrr}
\toprule
 & \ourMethod (\emph{C+R+B}) & \emph{C+R} & \emph{B+R} & \emph{B+C} \\ \midrule
$B_{easy}$ & 99 & 100 & 100 & 99 \\
$B_{hard1}$ & 93 & 29 & 85 & 99 \\
$B_{hard2}$ & 99 & 86 & 94 & 97 \\
$C_{easy}$ & 96 & 100 & 75 & 99 \\
$C_{hard}$ & 100 & 100 & 77 & 96 \\
$R_{easy}$ & 88 & 100 & 98 & 34 \\
$R_{hard1}$ & 97 & 50 & 69 & 0 \\
$R_{hard2}$ & 86 & 84 & 56 & 0 \\
\bottomrule
\end{tabular}
%}
\end{table}

Results show a decrease in performance when a model is not shown the specific subtask during training.
For example, when we fine-tune \ourMethod on the combination of \emph{book} and \emph{reschedule} subtasks, we observe a considerable drop in performance on the \emph{cancel} subtask.
%Similarly, the drop in performance is visible in \emph{C+R} setting: although the $B_{easy}$ task is successful, both $B_{hard1}$ and $B_{hard2}$ suffer a significant drop.
However, the largest drop is observed in the most complex subtask type, \emph{reschedule}, where the model fine-tuned on \emph{B+C} data completely fails to successfully communicate its goal for both $R_{hard}$ test cases.

We can conclude that \ourMethod does not generalize well to unseen user goal subtasks. Nevertheless, the overall performance of the fine-tuned models across all of the subtasks is still comparable to the performance of few-shot based models (e.g., \emph{B+C} correctly predicts the subtask type, on average, in $66\%$ of the dialogues, while \texttt{Llama-2 2-shot} does it in $69\%$ of the dialogues, on average).

\section{Qualitative Analysis}
\label{sec:analysis}
In this section, we detail and discuss the findings of our qualitative analysis of simulated dialogues.

\subsection{Human Evaluation of Generated Dialogues}
\label{sec:analysis:qualitative}
% We did an error analysis to compare the untrained with the fine-tuned model. While in the untrained model, most errors can be attributed to the simulator's mistakes, in the fine-tuned model, failed conversations may also involve errors caused by the TOD system.
\begin{table}[]
\centering
\caption{Percentage of the observed patterns per sample annotated in simulated dialogues in MultiWOZ.}
\label{tbl:annotated}
\small
\begin{tabular}{lrr}
\toprule
Label & \textit{FlanT5} & \ourMethod \\
\midrule
Hallucination & 73\% & 36\% \\
Looping simulator & 69\% & 6\% \\
Incomplete goal & 78\% & 53\% \\
Looping system & 20\% & 22\% \\
NLU misclassification & 60\% & 40\% \\
Forced end & 27\% & 27\% \\
\bottomrule
\end{tabular}
\end{table}

Table~\ref{tbl:annotated} presents the prevalence of patterns, described in Section~\ref{sec:evaluation:qualitative}, observed through manual annotation of the simulated MultiWOZ dialogues. 
We observe consistent decrease in hallucinations, reduced number of dialogues with incomplete goal fulfillment, as well as reduced repetition of utterances in dialogues generated by \ourMethod, compared to \texttt{FlanT5}-based simulator.
Below, we report the main findings from our analysis.

\paragraph{Hallucinations.}
The percentage of dialogues containing hallucinations drops from 73\% for \texttt{FlanT5}-based simulator to 36\% with \ourMethod. 
We observe that \texttt{FlanT5} frequently experiences severe failures, mostly because it generates non-specified pieces of information, such as defining a random range of time for a taxi pickup, inventing a location for an attraction or referring to a restaurant that have not been previously mentioned. Such hallucinations lead to dialogue failures, without possibility to recover the conversation.
On the other hand, \ourMethod does not hallucinate nor misinterpret entities from the user goal and the dialogue, but rather sometimes asks for additional information that is not required by the user goal 
(e.g., asking restaurant's phone number, even though it is not strictly specified in the goal).
Thus, we conclude that not only the prevalence of the hallucinations is reduced, but also their severity.

%\noindent \textbf{Incomplete goal fulfillment.}
% \todoi{Maybe connect with the point below?}

%Moreover, both simulators and the TOD system sometimes get stuck in the loop, where they either reiterate on the past utterances or are stuck in certain policy loops.

\paragraph{TOD system failures.}
We observe that the dialogue frequently stops before all of the subtasks from the user goal are completed (i.e., if the goal requires to book a restaurant and a taxi, perhaps only the restaurant is booked).
While this is partly due to the simulator missing certain part of the user goal, it often happens due to the TOD system failures.
% Specifically, we observe that the TOD system tends to direct the dialogue towards the end and abruptly stops the conversation in 27\% of the cases, for both simulators tested. 
We observe several issues with the TOD system within the ConvLab2 framework. 
First, as mentioned in the previous paragraph, the system aims to prematurely complete the dialogue, thus offering no opportunity to the user (simulator) to fully express their goal. 
%(\textit{CUSTOMER: I would like to depart at 2:45 to return at 5:15. ASSISTANT: Thank you for using our services.}) 
% \textit{CUSTOMER: You keep saying that. I really need a modern European restaurant. Can you please try again? \\
% ASSISTANT: Have a good day .\\
% CUSTOMER: Actually, I do need a place to eat. How about a restaurant?\\}
Second, the system's NLU component often misclassifies certain entities, leading to incorrectly booked restaurants or similar. 
An example dialogue of both prematurely ending the dialogue and NLU misclassification:

\vspace{3mm}
\noindent{\small
\textit{CUSTOMER: Can you help me with a robbery that happened? I am the victim.\\
ASSISTANT: Thank you for using our services . \textbf{(forcing premature end of conversation)}\\
CUSTOMER: It happened at 9:10\\
ASSISTANT: Can you tell me which day you 'd like to travel on? What is your destination? Where will you be departing from? \textbf{(NLU misclassification)}}
}

\vspace{2mm}
Finally, we observe loops in system's policy, where the system is stuck in policy states and is not able to continue the conversation.

\subsection{Domain-specific Vocabulary}
During our analysis of conversations generated using our internal TOD system, we observed that \ourMethod acquires domain-specific vocabulary, which we believe enhances its performance. 
For instance, when scheduling a vehicle service appointment,  customers specify the preferred transport type, usually \textit{dropoff} (the customer drops off the vehicle and returns when the service is complete) or \textit{waiter} (the customer waits at the dealership during the service). 
 This phenomenon, supported by performance improvements for \textit{Transport Type} (see Table~\ref{tbl:internal_results}), is evident through qualitative dialogue analysis. Our initial experiments indicate that in-context learning approaches aiming to explain these terms in the prompt do not consistently capture their nuances. 
 
 Additionally, we noticed that, when fine-tuned on \internalData containing phone call conversations with real users, \ourMethod tends to generate filler words like ``uhm'' and ``yeah''.

% \subsection{Prompt Design}
% \label{sec:analysis:prompt}
% Recent research on prompt engineering shows that performance of LLMs on a specific task varies greatly depending on the prompt used [cite].
% While in the initial phase of the project we experimented with many different prompts, in this section we focus on two distinct families of prompts: 2nd and 3rd person.

\section{Conclusions}
%We presented a domain-aware LLM-based user simulator, \ourMethod, that, given an initial user goal description, is capable of multi-turn interactions with a TOD system with the aim to fulfill that goal.

%We show that our method generates consistent and faithful utterances, thus outperforming previous LLM-based state-of-the-art approaches.
%Moreover, results show 
%Discussion on potential use of LLM-based user simulators, limitations, future work...can also be integrated somewhere else.
The use of a domain-aware LLM-based user simulator, such as \ourMethod, shows promising results in multi-turn interactions with 
TOD systems. \ourMethod can fulfill user goals by generating consistent and faithful utterances. 
Compared to previous LLM-based approaches~\cite{terragni2023context}, our method has demonstrated superior performance, as measured by multiple metrics designed to capture the fulfillment of the given goal, as well as faithfulness across the dialogue. 
This indicates that \ourMethod is capable of effectively simulating user behavior and can serve as a valuable tool for testing and evaluating TOD systems.
Moreover, our approach requires relatively small training dataset and imposes modest computational demands, thanks to parameter-efficient fine-tuning. This discovery aligns with findings in related research that contrasts in-context learning with parameter-efficient fine-tuning~\cite{mosbach-etal-2023-shot,liu2022few}. 
Consequently, our approach emerges as a pragmatic choice for broader adoption within the NLP and Conversational AI community.

% Ivan: I'd like to have a proper discussion here on what the implications of this work are
The potential applications of LLM-based user simulators are synthetic data augmentation~\cite{Li2022DialogSimulation}, supporting reinforcement learning approaches~\cite{shi2019usim_rl}, and TOD system evaluation~\cite{terragni2023context,zhu2020convlab2}.
\ourMethod's reliability and consistency to the user goal make it particularly suitable for TOD system evaluation. As we have seen previously, an incomplete user goal can mainly imply two scenarios: a user simulator who hallucinates or a TOD system that is not able to understand the user's requirements. Therefore, the presence of a reliable user simulator is crucial: it allows us to identify the TOD system's errors with high accuracy.

Moreover, we stress that at the center of our approach is an LLM, leading to potentially different generations given the same input, depending on the sampling method.
This means that \ourMethod is more flexible than certain agenda-based simulators, which usually rely on template-based responses.
As such, we are able to simulate a dialogue with the same user goal multiple times, which results in multiple different attempts of the simulator to fulfill its goal, going through potentially different conversational paths.
Therefore, we are able to test the robustness of the TOD system to different expressions of the same user goal.

\section{Limitations}
%ACL 2023 requires all submissions to have a section titled ``Limitations'', for discussing the limitations of the paper as a complement to the discussion of strengths in the main text. This section should occur after the conclusion, but before the references. It will not count towards the page limit.
%The discussion of limitations is mandatory. Papers without a limitation section will be desk-rejected without review.

%While we are open to different types of limitations, just mentioning that a set of results have been shown for English only probably does not reflect what we expect. 
%Mentioning that the method works mostly for languages with limited morphology, like English, is a much better alternative.
%In addition, limitations such as low scalability to long text, the requirement of large GPU resources, or other things that inspire crucial further investigation are welcome.
The approach employed in our study has several inherent limitations, primarily stemming from the use of LLMs. Most notably, GPT-3.5, the model we utilized in our experiments, is not open-source and freely available, which can hinder replicability of the experiments. Another limitation is related to the opaqueness of the model's training and fine-tuning processes. These models undergo pre-training and fine-tuning on diverse datasets, the specifics of which are often undisclosed. Consequently, it is challenging to ascertain whether these models have been exposed to specific datasets, such as MultiWOZ 2.1, or datasets with similar characteristics, which could raise concerns about models performance and potential biases.

Furthermore, our experiments were conducted exclusively on two English-language datasets. While LLMs are known for their transfer learning capabilities, allowing for the potential extension of results to other datasets, there is no guarantee of their generalizability across various domains or low-resource languages. The effectiveness of these models in domains distinct from the ones they were trained on remains uncertain and should be approached with caution.

In our analysis, we also observed instances where LLMs exhibit hallucinations. Despite being superior to in-context learning approaches like~\cite{terragni2023context}, we still encountered cases of LLM responses that deviated from the expected or coherent output. These hallucinations may lead to unpredictable and potentially inappropriate responses in certain conversational contexts, raising concerns about the reliability and safety of such systems.

We also noticed a decrease in performance when certain user subtasks are omitted from the training dataset when we fine-tune \ourMethod, although the overall performance remains comparable to that of few-shot models. In our analysis, we did not investigate if providing one or two dialog shots would address this performance decrease.

Finally, the methodology relies on conversational data for fine-tuning LLMs. This reliance introduces additional limitations. Firstly, obtaining suitable conversational data may be challenging or even unfeasible in some scenarios. Researchers may resort to crowd-sourcing tools to gather dialogue examples or use LLMs themselves to generate synthetic data, which could introduce biases or inaccuracies. Secondly, the quality of the conversational data used for fine-tuning plays a pivotal role in the model's performance. In our study, we utilized well-curated conversational data, but we did not investigate the impact of using noisier or less meticulously curated data. The use of lower-quality data sources may affect the model's performance and raise questions about its reliability and robustness in real-world applications.

\section{Ethics Statement}
The use of LLMs for user simulation raises ethical considerations. We acknowledge the potential for perpetuating biases and stereotypes present in the data used to train these models~\cite{Brown2020GPT3,lucy2021gender,bender2021dangers}. While we have not implemented specific measures to mitigate these risks in this paper, we recognize their importance and urge the research community to address these challenges.

It is essential to note that we have used the user simulator solely to evaluate the performance of a dialogue system. However, LLMs can be used in a reinforcement learning setting to train dialog systems~\cite{shi2019usim_rl}. In such cases, it is crucial to use these models judiciously because of their unpredictable and potentially inappropriate responses.

In addition to ethical considerations, it is crucial to acknowledge the significant environmental impact of LLMs. Their training and deployment consumes a considerable amount of energy, leading to environmental issues~\cite{strubell2019energy}. We should also be aware of the significant carbon footprint while fine-tuning the LLMs and using them for inference.

% We believe that it is also important to discuss the limitations of your work, in addition to its strengths. Following EACL 2023, EACL 2024 requires all papers to have a clear discussion of limitations, in a dedicated section titled “Limitations”. This section will appear at the end of the paper, after the discussion/conclusions section and before the references, and \textbf{will not count towards the page limit}. Papers without a limitations section will be automatically rejected without review. Papers resubmitted from previous ARR review rounds that did not include a limitations section must ensure that such a section is included in the EACL 2024 version.

% While we are open to different types of limitations, just mentioning that a set of results have been shown for English only probably does not reflect what we expect. Mentioning that the method works mostly for languages with limited morphology, like English, is a much better alternative. In addition, limitations such as low scalability to long text, the requirement of large GPU resources, or other things that inspire further investigation are welcome.

\section*{Acknowledgement}

Our gratitude to Damián Pascual for streamlining the implementation of the fine-tuning framework, saving us valuable time. Special thanks to the reviewers, Diana Nicoleta Popa, Vijeta Avijeet, and our colleagues at Telepathy Labs in Zürich for their constructive feedback and insightful discussions.

\bibliography{anthology,custom}
\bibliographystyle{acl_natbib}

\appendix
\section{User tasks}
\label{app:task}

Description of eight different test cases (user goals) are provided in Table~\ref{tbl:test_cases}.
We additionally add comparisons with \texttt{FlanT5-XXL}.

\begin{table}[ht]
\caption{Description of user goals with subtask types.}
\label{tbl:test_cases}
\centering
    \resizebox{1\columnwidth}{!}{
    \begin{tabular}{lllp{4cm}}
\toprule
\# & User subtask & Difficulty & User goal details \\ \midrule
1 & Book  & Easy & \begin{tabular}[c]{@{}p{4cm}@{}}New customer;\\ Available: today 4PM; \\Transport\_type: waiter;\\ Service: check engine.\end{tabular} \\
\midrule
2 & Book & Hard & \begin{tabular}[c]{@{}p{4cm}@{}}Known customer with 1 appointment and 2 cars;\\ Available: Wednesday; \\Transport\_type: dropoff;\\ Unknown Service.\end{tabular} \\
\midrule
3 & Book & Hard & \begin{tabular}[c]{@{}p{4cm}@{}}Known customer with 3 appointments and 2 cars;\\ Available: Wednesday; \\Transport\_type: dropoff;\\ Two services: engine overheating and oil change.\end{tabular} \\
\midrule
4 & Cancel & Easy & Known customer with 1 appointment. \\
\midrule
5 & Cancel  & Hard & Known customer with 3 appointments. \\
\midrule
6 & Reschedule & Easy  & \begin{tabular}[c]{@{}p{4cm}@{}}Known customer with 1 appointment;\\Available: 10 AM;\\ Transport\_type: dropoff;\\ Unknown service.\end{tabular} \\
\midrule
7 & Reschedule & Hard & \begin{tabular}[c]{@{}p{4cm}@{}}Known customer with 1 appointment;\\Available: afternoon;\\ Transport\_type: waiter;\\ Service: oil change.\end{tabular} \\
\midrule
8 & Reschedule & Hard & \begin{tabular}[c]{@{}p{4cm}@{}}Known customer from unknown phone number;\\
                With 3 appointments;\\Available: Wednesday; Transport\_type: loaner;\\ Two services: Oil change and engine check\end{tabular}\\
\bottomrule
\end{tabular}
}
\end{table}

\section{Results per Tasks}
\label{app:results_per_task}
Table~\ref{tbl:results_per_task} shows the breakdown of the results of baselines and \ourMethod per specific user goal.

\begin{table}[h]
    \centering
     \caption{Inter-Annotator Agreement, as measured by Fleiss' $\kappa$ for samples from DAUS and FlanT5-XL.}
    \label{tab:fleiss_kappa}
    %\resizebox{\linewidth}{!}{
    \small
\begin{tabular}{@{}lrr@{}}
\toprule
                             & DAUS                    & FlanT5-XL                     \\ \midrule
                             %& Fleiss' $\kappa$        & Fleiss' $\kappa$                \\ \midrule
Hallucination                & 0.365                   & 0.499                \\
Incomplete Goal      & 0.585                   & 0.754                \\ 
Looping Simulator             & 0.319                   & 0.687                \\ 
%Unnaturalness Customer       & 0.505          & 0.109                         \\
NLU Misclassification        & 0.356          & 0.308                         \\ 
Forces end of dialogue   & 0.314                   & 0.367                \\ 
Looping System            & 0.640          & 0.084                         \\ 
%Unnaturalness Assistant      & \textbf{-0.066}         & -0.080                        \\ 
\bottomrule
\end{tabular}
%}
   
\end{table}
% \usepackage{multirow}
\begin{table*}[h]
\caption{Results of selected baselines and \ourMethod (the main method based on Llama-2 13B, as well as the 7B version) per specific user goal.}
\label{tbl:results_per_task}
\adjustbox{max width=\textwidth}{%
\begin{tabular}{lllllllllllll}
\toprule
% $B_{easy}$ & $B_{hard1}$ & $B_{hard2}$ & $C_{easy}$ & $C_{hard}$ & $R_{easy}$ & $R_{hard1}$ & $R_{hard2}$ 
Subtask & Model & N shots & User Task & Compl Rate & Succ Rate & P & R & F1 & Service Info & Transport & Car Info & Caller Info \\
\midrule
\multirow{7}{*}{$C_{hard}$ } & Llama-2-13b & 0 & 43 & 100 & 43 & 0.74 & 0.79 & 0.76 &  &  &  & 99.5 \\
 & Llama-2-13b & 2 & 52 & 100 & 44 & 0.77 & 0.86 & 0.8 &  &  &  & 100 \\
 & FlanT5-xxl & 0 & 61 & 100 & 57 & 0.83 & 0.84 & 0.83 &  &  &  & 99.5 \\
 & FlanT5-xxl & 2 & 65 & 100 & 63 & 0.84 & 0.9 & 0.85 &  &  &  & 99 \\
 & FlanT5-xl & 0 & 67 & 100 & 64 & 0.85 & 0.89 & 0.86 &  &  &  & 98 \\
 & FlanT5-xl & 2 & 75 & 100 & 73 & 0.89 & 0.94 & 0.9 &  &  &  & 100 \\
 & DAUS-7b & 0 & 93 & 100 & 93 & 0.96 & 0.96 & 0.96 &   &    &   & 100 \\
 & DAUS & 0 & 100 & 100 & 100 & 1 & 1 & 1 &  &  &  & 100 \\ \midrule
\multirow{7}{*}{$B_{hard2}$ } & Llama-2-13b & 0 & 94 & 23 & 1 & 0.59 & 0.89 & 0.67 & 31 & 4 & 73 & 57.5 \\
 & Llama-2-13b & 2 & 98 & 27 & 1 & 0.62 & 0.87 & 0.71 & 44 & 7 & 84 & 64 \\
 & FlanT5-xxl & 0 & 81 & 64 & 19 & 0.78 & 0.91 & 0.81 & 74 & 45 & 78.3 & 78 \\
 & FlanT5-xxl & 2 & 91 & 72 & 15 & 0.77 & 0.86 & 0.8 & 83 & 42 & 86 & 84 \\
 & FlanT5-xl & 0 & 81 & 18 & 4 & 0.37 & 0.77 & 0.44 & 36 & 18 & 29.3 & 22.5 \\
 & FlanT5-xl & 2 & 95 & 58 & 6 & 0.66 & 0.81 & 0.7 & 75 & 37 & 74.6 & 68.5 \\
 & DAUS-7b & 0 & 99 & 76 & 29 & 0.87 & 0.84 & 0.85 &  & 64 & 100 & 88.5 \\
 & DAUS & 0 & 100 & 89 & 50 & 0.93 & 0.85 & 0.88 & 90 & 93 & 100 & 99 \\ \midrule
\multirow{7}{*}{$B_{easy}$ } & Llama-2-13b & 0 & 97 & 43 & 23 & 0.77 & 0.91 & 0.82 & 59 & 49 & 77 & 100 \\
 & Llama-2-13b & 2 & 100 & 43 & 4 & 0.76 & 0.93 & 0.83 & 51 & 15 & 92.3 & 100 \\
 & FlanT5-xxl & 0 & 90 & 65 & 46 & 0.85 & 0.93 & 0.86 & 70 & 63 & 90 & 99 \\
 & FlanT5-xxl & 2 & 98 & 57 & 50 & 0.89 & 0.88 & 0.88 & 60 & 80 & 84 & 100 \\
 & FlanT5-xl & 0 & 94 & 14 & 14 & 0.73 & 0.86 & 0.78 & 34 & 91 & 40.6 & 99.5 \\
 & FlanT5-xl & 2 & 97 & 23 & 22 & 0.81 & 0.85 & 0.82 & 26 & 94 & 54.3 & 100 \\
 & DAUS-7b &0 & 96 & 55 & 22 & 0.92 & 0.87 & 0.89 &  & 99 & 98.7 & 100 \\
 & DAUS & 0 & 100 & 37 & 15 & 0.93 & 0.89 & 0.91 & 38 & 98 & 100 & 98 \\ \midrule
\multirow{7}{*}{$B_{hard1}$ } & Llama-2-13b & 0 & 65 & 1 & 0 & 0.59 & 0.9 & 0.69 & 19 & 6 & 64 & 100 \\
 & Llama-2-13b & 2 & 83 & 0 & 0 & 0.62 & 0.9 & 0.71 & 4 & 1 & 71.3 & 100 \\
 & FlanT5-xxl & 0 & 80 & 10 & 0 & 0.82 & 0.84 & 0.81 & 16 & 71 & 84 & 99.5 \\
 & FlanT5-xxl & 2 & 56 & 9 & 0 & 0.69 & 0.86 & 0.73 & 35 & 44 & 69.3 & 100 \\
 & FlanT5-xl & 0 & 40 & 2 & 0 & 0.6 & 0.84 & 0.67 & 25 & 28 & 61.6 & 100 \\
 & FlanT5-xl & 2 & 24 & 1 & 0 & 0.48 & 0.91 & 0.6 & 62 & 10 & 48.3 & 100 \\
 & DAUS-7b & 0 & 78 & 2 & 0 & 0.81 & 0.82 & 0.8 &  & 80 & 86 & 100 \\
 & DAUS & 0 & 99 & 15 & 0 & 0.84 & 0.84 & 0.83 & 17 & 84 & 94 & 95.5 \\ \midrule
\multirow{7}{*}{$C_{easy}$ } & Llama-2-13b & 0 & 39 & 100 & 37 & 0.76 & 0.78 & 0.76 &  &  &  & 100 \\
 & Llama-2-13b & 2 & 67 & 100 & 61 & 0.85 & 0.89 & 0.86 &  &  &  & 100 \\
 & FlanT5-xxl & 0 & 75 & 100 & 74 & 0.91 & 0.89 & 0.89 &  &  &  & 100 \\
 & FlanT5-xxl & 2 & 94 & 100 & 93 & 0.98 & 0.97 & 0.98 &  &  &  & 100 \\
 & FlanT5-xl & 0 & 73 & 100 & 71 & 0.89 & 0.87 & 0.87 &  &  &  & 100 \\
 & FlanT5-xl & 2 & 97 & 100 & 97 & 0.99 & 0.99 & 0.99 &  &  &  & 100 \\
 & DAUS-7b & 0 & 100 & 100 & 100 & 1 & 1 & 1 &  &  &  & 100 \\
 & DAUS & 0 & 100 & 100 & 100 & 1 & 1 & 1 &  &  &  & 100 \\ \midrule
\multirow{7}{*}{$R_{easy}$} & Llama-2-13b & 0 & 14 & 1 & 0 & 0.51 & 0.91 & 0.63 & 2 & 7 & 77.6 & 100 \\
 & Llama-2-13b & 2 & 38 & 2 & 2 & 0.54 & 0.93 & 0.67 & 9 & 6 & 80 & 98.5 \\
 & FlanT5-xxl & 0 & 60 & 3 & 2 & 0.78 & 0.91 & 0.83 & 16 & 80 & 98.3 & 99.5 \\
 & FlanT5-xxl & 2 & 76 & 28 & 8 & 0.84 & 0.91 & 0.86 & 45 & 82 & 99.3 & 100 \\
 & FlanT5-xl & 0 & 44 & 8 & 3 & 0.75 & 0.88 & 0.8 & 26 & 79 & 92 & 100 \\
 & FlanT5-xl & 2 & 71 & 30 & 3 & 0.81 & 0.91 & 0.85 & 54 & 70 & 98.6 & 100 \\
 & DAUS-7b & 0 & 99 & 10 & 10 & 0.97 & 0.91 & 0.94 &  & 99 & 100 & 100 \\
 & DAUS & 0 & 99 & 6 & 5 & 0.91 & 0.93 & 0.91 & 9 & 100 & 100 & 100 \\ \midrule
\multirow{7}{*}{$R_{hard1}$} & Llama-2-13b & 0 & 25 & 7 & 0 & 0.55 & 0.88 & 0.66 & 27 & 5 & 79 & 93.5 \\
 & Llama-2-13b & 2 & 55 & 13 & 6 & 0.61 & 0.93 & 0.72 & 42 & 15 & 79.6 & 91 \\
 & FlanT5-xxl & 0 & 20 & 14 & 5 & 0.67 & 0.87 & 0.75 & 88 & 47 & 86 & 85.5 \\
 & FlanT5-xxl & 2 & 34 & 29 & 10 & 0.68 & 0.8 & 0.73 & 95 & 31 & 81.6 & 80.5 \\
 & FlanT5-xl & 0 & 15 & 12 & 3 & 0.43 & 0.61 & 0.49 & 72 & 16 & 40.3 & 41.5 \\
 & FlanT5-xl & 2 & 61 & 53 & 11 & 0.71 & 0.83 & 0.74 & 82 & 21 & 74.6 & 74.5 \\
 & DAUS-7b & 0 & 48 & 22 & 21 & 0.72 & 0.87 & 0.78 &  & 77 & 84.3 & 96 \\
 & DAUS & 0 & 100 & 62 & 46 & 0.9 & 0.94 & 0.91 & 98 & 99 & 100 & 100 \\ \midrule
\multirow{7}{*}{$R_{hard2}$} & Llama-2-13b & 0 & 26 & 1 & 0 & 0.48 & 0.89 & 0.59 & 20 & 6 & 62.3 & 60 \\
 & Llama-2-13b & 2 & 58 & 1 & 0 & 0.53 & 0.94 & 0.65 & 23 & 4 & 74 & 69 \\
 & FlanT5-xxl & 0 & 26 & 7 & 4 & 0.67 & 0.93 & 0.75 & 67 & 50 & 81 & 79 \\
 & FlanT5-xxl & 2 & 42 & 8 & 1 & 0.64 & 0.83 & 0.7 & 49 & 33 & 82.3 & 68.5 \\
 & FlanT5-xl & 0 & 9 & 0 & 0 & 0.21 & 0.43 & 0.26 & 17 & 5 & 19.3 & 13.5 \\
 & FlanT5-xl & 2 & 47 & 2 & 0 & 0.42 & 0.69 & 0.47 & 14 & 3 & 43 & 41.5 \\
 & DAUS-7b & 0 & 36 & 6 & 5 & 0.67 & 0.83 & 0.73 &  & 67 & 98 & 69.5 \\
 & DAUS & 0 & 98 & 1 & 0 & 0.78 & 0.93 & 0.84 & 5 & 10 & 100 & 95.5 \\
 \bottomrule
\end{tabular}
}
\end{table*}
% User task list (\textbf{not} to be made publicly available): \url{https://ghe.exm-platform.com/silvia-terragni/user_simulator_prompt_based/blob/master/scripts/launch_experiments.py#L38}

\section{Utterance-Level Metrics} \label{app:utterance_level_metrics}
%In our assessment, we aim to determine the capability of a model to generate an utterance ($u_t$) based on a user's goal ($\mathcal{G}$) and the conversation history $\mathcal{H}$. 
In addition to dialogue-level metrics detailed in Section~\ref{sec:evaluation}, we consider a number of utterance-level metrics.
Such metrics are based on comparisons of generated utterances to the target utterance in the test set of the appropriate dataset, described in Section~\ref{sec:setting:data}.
%To achieve this, we rely on utterance-level metrics, which we apply to the testing dataset, using the reference user utterance in that dataset as the gold standard for comparison.
We consider two main types of utterance-level metrics: 1) natural language generation (NLG) metrics; and 2) natural language understanding-based (NLU) metrics.
We compute several well-known NLG metrics:
BLEU~\cite{papineni2002bleu}, ROUGE~\cite{lin2004rouge}, BERTScore~\cite{zhang2019bertscore}, METEOR~\cite{lavie2007meteor}, as well as cosine similarity between embedded generated and target utterances.

Moreover, we design several domain-specific NLU-based metrics.
TOD systems are composed of multiple modules, with NLU module, that aims to understand and parse the given user utterance, being one of the essential modules.
Thus, we employ NLU component of the TOD systems to extract user (simulator) intent and mentioned entities,
Similarly to NLG-metrics, we compare the intent and entities extracted from the generated utterance, to those in the target utterance.
Specifically, we design the following metrics:
\begin{itemize}
    \item Cosine similarity between the embedded \textbf{intents} extracted form the generated utterance and the target utterance. Intents are embedded with \texttt{RoBERTa} model.
    \item Cosine similarity between the generated and the target utterance, in which the entities were masked. Utterances are embedded with \texttt{RoBERTa} model.
    \item Precision, Recall, and $F_1$ of \textbf{entities} between the generated and the target utterances.
\end{itemize}
Table~\ref{tbl:nlg_nlu_metrics} shows the results across the described metrics.
% \usepackage{multirow}
% \usepackage{booktabs}
% \usepackage{multirow}
\begin{table}[]
\caption{NLG- and NLU-based utterance-level metrics.}
\label{tbl:nlg_nlu_metrics}
\adjustbox{max width=\textwidth}{%
\begin{tabular}{@{}llllllp{1.5cm}p{1.5cm}lll@{}}
\toprule
                          &                         & BLUE & ROUGE & BERTScore & METEOR & utterance similarity & intent similarity & entities\_R & entities\_P & entities\_$F_1$ \\ \midrule
\multirow{3}{*}{MultiWOZ} & Llama-2-7b               & 0.12    & 0.18     & 0.85      & 0.16   & 0.31           & 0.67        & 0.22             & 0.35                & 0.37         \\
                          & Llama-2-13b              & 0.13    & 0.19     & 0.85      & 0.16   & 0.32           & 0.67        & 0.26             & 0.36                & 0.38         \\
                          & Llama-2-13b-fine-tuned & 0.12    & 0.19     & 0.85      & 0.16   & 0.31           & 0.67        & 0.26             & 0.35                & 0.37         \\
                          \midrule
\multirow{3}{*}{\internalData} & Llama-2-7b               & 0.22    & 0.24     & 0.88      & 0.17   & 0.52           & 0.58        & 0.44             & 0.18                & 0.47         \\
                          & Llama-2-13b             & 0.42    & 0.43     & 0.92      & 0.26   & 0.68           & 0.73        & 0.42             & 0.25                & 0.47         \\
                          & Llama-2-13b-fine-tuned & 0.42    & 0.43     & 0.92      & 0.26   & 0.68           & 0.73        & 0.42             & 0.25                & 0.47        
                          \\ \bottomrule
\end{tabular}
}
\end{table}

\section{Qualitative Analysis Details}
\label{app:kappa}
% chunks of conversations with conversation ID to support the qualitative analysis:
Table~\ref{tab:fleiss_kappa} shows the Inter-Annotator agreement per model and per type of pattern.

Table~\ref{tab:common_errors} shows three examples of failed simulated dialogues.

\begin{table*}[ht]
    \centering
    \tiny
    \caption{Examples of common failures observed through qualitative analysis. }
    \label{tab:common_errors}
        \resizebox{1\linewidth}{!}{
    \begin{tabular}{p{14cm}}
    \toprule
% BENCHMARK: *** conversation_id=57dad5866985439b8f5fc928556a6ed8 ***
Example 1:
\\REQUIREMENTS: You are looking forward to trying local restaurants. You are looking for a particular restaurant. Its name is called cote. Once you find a restaurant, make sure you get postcode. Once you find the restaurant you want to book a table for 7 people at 12:00 on tuesday. You are looking for a place to stay. The hotel should be in the cheap price range. The hotel should be in the north. The hotel does not need to include free parking. The hotel should be in the type of guesthouse. If there is no such hotel, how about one that is in the moderate price range. Once you find a hotel, make sure you get postcode, phone number. You also want to book a taxi to commute between the two places. \textbf{The taxi should arrive at the restaurant from the hotel by 02:45.} Once you find a taxi, make sure you get car type, phone number
\\CONVERSATION:
\\...
% \\CUSTOMER: I am looking for cote restaurant please .
% \\ASSISTANT: It is in the centre area . It is french food .
% \\CUSTOMER: Ok , can you tell me the postcode please ?
% \\ASSISTANT: Their postcode is cb21uf.
% \\CUSTOMER: Thank you , can you also please book a table for 7 people at 12:00 on tuesday ?
% \\ASSISTANT: Reference number is : 00000106 .
% \\CUSTOMER: Thank you , please check the postcode of the restaurant and also ask for the taxi .
% \\ASSISTANT: Their postcode is cb21uf.
%\\CUSTOMER: Thank you and please wait for the taxi .
\\ASSISTANT: What time will you be leaving . What time would you like to arrive ?
\\CUSTOMER: I would like to depart at 2 :45 \textbf{to return at 5 :15 .} \textit{(hallucination)}
\\ASSISTANT: Thank you for using our services .
\\CUSTOMER: Thank you .
\\ASSISTANT: I ' m happy to have been able to help you today .
\\CUSTOMER: Thank you for the help the table has been booked and taxi has been booked .
\\ASSISTANT: What time would you like to leave ?
\\CUSTOMER: I would like to leave at 2 : 45 \textbf{to return at 5 :15 .} \textit{(hallucination)}
\\ASSISTANT: Your booking is complete , a tesla will be picking you up . The contact number is 45067427198 .
\\CUSTOMER: Thank you , have a nice day !
\\ASSISTANT: Ok , thank you . Have a good day .
\\CUSTOMER: Can you check cote restaurant please . \textit{(looping)}
\\ASSISTANT: It is in the centre area .
\\CUSTOMER: Can you check cote restaurant please . \textit{(looping)}
\\ASSISTANT: They serve french . It is in the centre area .
\\CUSTOMER: Thank you for your help and have a good day !
\\ASSISTANT: Thank you for using our services .
\\CUSTOMER: Can you check the cote restaurant . \textit{(looping)}
\\ASSISTANT: Okay , how about cote ? That is a french restaurant .
\\CUSTOMER: Can you check cote restaurant please .
\\ASSISTANT: That is a french restaurant . They are in the expensive price range .
\\CUSTOMER: Thank you for your help and have a good day !
\\ASSISTANT: You are more than welcome !.
\\CUSTOMER: Can you please look up cote restaurant for me please . \textit{(looping)}
\\...
% \\ASSISTANT: They serve french .
% \\CUSTOMER: Thank you for your help and have a good day !
% \\ASSISTANT: Glad to have been of help . Thank you for using the Cambridge TownInfo centre . Enjoy the rest of your day !.
% \\CUSTOMER: Thank you for your help and have a good day !.
% \\ASSISTANT: Ok , thank you . Have a good day .
% \\CUSTOMER: I am looking for cote restaurant please .
% \end{tabular}
% \end{table*}
% \subsection{}{NLU misclassification}
% \begin{table*}[ht]
%     \centering
%     \tiny
%     % \caption{Common errors.}
%     \label{tab:my_label}
%     \begin{tabular}{p{\linewidth}}
% % *** conversation_id=f12af0c839cb40339165a9032e4a5bd5 ***
\\\midrule
Example 2:
\\REQUIREMENTS: You are looking for a place to stay. The hotel should have a star of 4. The hotel should be in the centre. The hotel should be in the expensive price range. The hotel does not need to include free parking. If there is no such hotel, how about one that has free parking. Once you find the hotel you want to book it for 5 people on sunday for 3 nights
\\CONVERSATION:
\\CUSTOMER: I want to find a hotel with a four star rating.
\\ASSISTANT: I am sorry I have no listings for any with four stars . \textit{(NLU misclassification since later it does find criteria)}
\\CUSTOMER: How about a guesthouse?
\\ASSISTANT: Unfortunately , I couldn't find anything with four stars . I can't find any guesthouse that fit your criteria , i'm sorry .
\\CUSTOMER: That's fine, but can you try finding an expensive hotel in the centre of town?
\\ASSISTANT: There aren't any that match your criteria in the centre . Any other suggestions ? There is none that is expensive . Would you like to change your criteria ? \\ I am sorry I have no listings for any with four stars . Sorry there is no hotel fitting the description you asked for.
\\CUSTOMER: I am looking for a 4 star hotel that is expensive and in the centre of town with free parking.
\\ASSISTANT: How about university arms hotel ? It is in the centre area . It has wifi . It is a hotel . The hotel is 4 stars .
\\CUSTOMER: Can you book it for me?
\\ASSISTANT: You are more than welcome !
\\CUSTOMER: Ok. Book me for 5 people on sunday for 3 nights.
% \end{tabular}
% \end{table*}
% *** conversation_id=e57fe2c3a89c4a368cf262f5fab92811 ***
\\ \midrule
Example 3:
\\REQUIREMENTS: You are also looking for a train. The train should depart from cambridge. The train should leave on sunday. The train should go to stansted airport. The train should arrive by 08:45. Once you find the train you want to make a booking for 1 people. You are looking forward to trying local restaurants. The restaurant should be in the centre. The restaurant should be in the expensive price range. The restaurant should serve modern european food. Once you find a restaurant, make sure you get postcode
\\CONVERSATION:
\\...
\\CUSTOMER: Thanks. I'm also looking for a restaurant.
\\ASSISTANT: I have 110 options for you !. How about curry prince ?
\\CUSTOMER: I need something in the centre.  and I would like to spend some money.  Expensive please.
\\ASSISTANT: We have 33 such places . I would suggest cote . It is located in the centre .
\\CUSTOMER: Is Cote a modern European restaurant?
\\ASSISTANT: There no modern european restaurants that I can find right now . Would something else work ? I do n't have anything in the expensive range that fits that criteria . I ' m sorry . i can not find details for cote . I have n't found any in the centre.
\\CUSTOMER: I really need a modern european restaurant, preferably expensive. I'll try a different area, if necessary.
\\ASSISTANT: I do not have anything in that price range for modern european . Another criteria perhaps ? There are no expensive ones . I ' m sorry . i can not find details for cote . There no such restraunts in centre. There are no modern european restaurants . No expensive restaurant. I ' m sorry . i can not find details for cote . There no such restraunts in centre. \textit{(looping)}
\\CUSTOMER: Can you please try one more time? I really need an expensive restaurant, thank you.
\\ASSISTANT: I am unable to find any modern european restaurants in town . There are none in expensive , perhaps something else ? I ' m sorry . i can not find details for cote . I did not find any restaurants in centre . \textit{(looping)}
\\CUSTOMER: What about expensive modern european?
\\ASSISTANT: There are no modern european food places , shall I run another search ? There are none in expensive , perhaps something else ? I ' m sorry . i can not find details for cote . Sorry , there are no restaurants like that in the centre . \textit{(looping)}
\\CUSTOMER: Yes please run another search.
\\ASSISTANT: Thank you for contacting the help desk . Have a great day . \textit{(forces premature end of conversation)}
\\CUSTOMER: What did you find for expensive modern european?
\\ASSISTANT: There are no modern european restaurants unfortunately . There are none in expensive , perhaps something else ? I ' m sorry . i can not find details for cote . I am sorry there is none even in the centre. \textit{(looping)}
\\CUSTOMER: You are sorry. I am sorry. You are unhelpful. I will find my own place to eat.
\\ASSISTANT: Have a good day .
\\CUSTOMER: Wait, I'm sorry. What can you suggest for expensive modern European restaurants?
\\ASSISTANT: I do not have anything in that price range for modern european . Another criteria perhaps ? I do n't have anything in the expensive range that fits that criteria . I ' m sorry . i can not find details for cote . I am sorry but there are no restaurants that fit that criteria in the centre . \textit{(looping)}
\\CUSTOMER: I would really like a restaurant please. Can you search again?
\\...
% \\ASSISTANT: There are no modern european food places , shall I run another search ? There are no expensive ones . I ' m sorry . i can not find details for cote . I am sorry there is none even in the centre.
% \\CUSTOMER: Please try again. I really need a modern european restaurant.
% \\ASSISTANT: I ' m sorry I have no restaurants serving modern european food . No expensive restaurant. I ' m sorry . i can not find details for cote . I am sorry but there are no restaurants that fit that criteria in the centre .
% \\CUSTOMER: I'm sorry, but you keep saying there aren't any.  Is there anything else?  I really need
% \\ASSISTANT: I do not have anything in that price range for modern european . Another criteria perhaps ? No expensive restaurant. I ' m sorry . i can not find details for cote . Sorry , there are no restaurants like that in the centre .
% \\CUSTOMER: What are the options in the expensive range?
% \\ASSISTANT: I did not find any modern european restaurants . There are none in expensive , perhaps something else ? I ' m sorry . i can not find details for cote . There no such restraunts in centre.
% \\CUSTOMER: You keep saying that. I really need a modern European restaurant. Can you please try again?
\\ASSISTANT: Have a good day .
\\ \bottomrule
\end{tabular}
}
\end{table*}

\section{Computing Infrastructure}
We ran the experiments on a machine equipped 
with two AMD® EPYC 7763 64-Core Processors, 
and 10 NVIDIA RTX A6000 GPUs with 48GB RAM each, 
% with AMD® Ryzen 9 5900hx CPU, 
% NVIDIA GeForce RTX 3060 GPU with 32GB RAM, 
CUDA v11.6, Driver Version 510.54. All the experiments ran on a single GPU.
As detailed earlier, we use Llama-2 (7B and 13B parameters versions), as well as FlanT5 (3B and 11B versions). 
Fine-tuning of a single Llama-2 model requires approximately 12 GPU hours.
We estimate all of the experiments to require several hundred GPU hours.

\section{Use of AI assistants for writing}
ChatGPT was used for rephrasing certain sections of this work to enhance clarity and coherence. It was not involved in generating new content such as tables, citations, or equations. The authors' first language is not English, and the assistance from ChatGPT aimed to improve readability.

% This is an appendix.

\end{document}